\documentclass[]{youtu}

\usepackage{amsmath,amssymb,amsfonts}
\usepackage{booktabs}
\usepackage{multirow}
\usepackage{array}
\usepackage{tabularx}
\usepackage{graphicx}
\usepackage{xspace}
\usepackage{microtype}
\usepackage[T1]{fontenc}
\usepackage[utf8]{inputenc}
\usepackage{inconsolata}
\usepackage{mathpazo}
\usepackage{enumitem}

\newcommand{\method}{\textsc{H3-World}\xspace}
\newcolumntype{Y}{>{\raggedright\arraybackslash}X}
\newcolumntype{C}[1]{>{\centering\arraybackslash}p{#1}}

\title{H3-World: Turning Language Understanding into World Control}

\author{%
    Danze Chen$^{1,2,\clubsuit}$
    \quad Zeqing Wang$^{1,2,\clubsuit}$
    \quad Ziyue Lin$^{3}$
    \quad Xingyi Yang$^{3,*}$
    \quad Yeying Jin$^{1,2,*,\diamondsuit}$
}
\affiliation{%
    $^{1}$ Tencent
    \quad $^{2}$ National University of Singapore
    \quad $^{3}$ The Hong Kong Polytechnic University
}
\project{https://danzer1xxxxchan.github.io/H3-World}
\sourcecode{https://github.com/Danzer1xxxxChan/H3-World}
\model{https://huggingface.co/DANNY621/H3-World}

\teaser{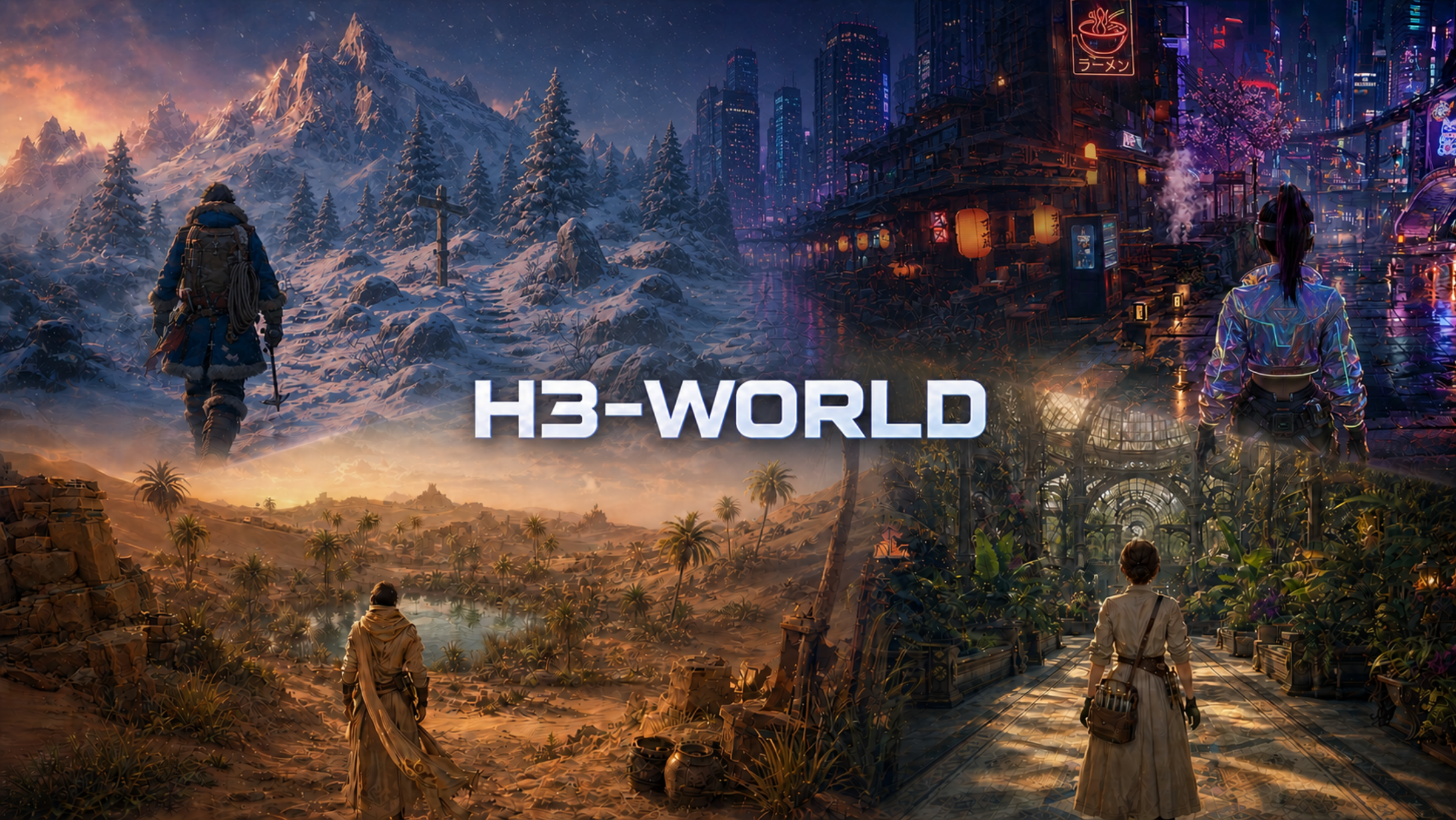}{\method enables character and camera control across diverse visual environments.}{fig_teaser}

\hypersetup{
    pdftitle={H3-World: Turning Language Understanding into World Control},
    pdfauthor={Danze Chen, Zeqing Wang, Ziyue Lin, Xingyi Yang, Yeying Jin},
    pdfsubject={Interactive World Models},
    pdfkeywords={World Models, Interactive Video Generation, Action Conditioning, Video Diffusion Models, Parameter-Efficient Adaptation}
}

\begin{document}

\abstract{
We present \textbf{\method}, an efficient framework that turns the 33B MiniMax-H3 video generator into an interactive world model. Our key finding is that, as large video generators become more capable, language is emerging as a natural interface for control. MiniMax-H3, for example, already supports zero-shot control of character behavior and camera motion through natural-language instructions. Building on this, \method turns this coarse language interface into precise, temporally grounded world control, without introducing dedicated action modules. Specifically, we represent each action as a structured combination of character and camera instructions, and align them with the corresponding temporal video latents. To make the control temporally precise, we further introduce temporal attention routing, which restricts each instruction to its intended time interval and reduces control leakage across actions. Importantly, \method directly reuses the semantic representations learned during large-scale video pretraining and requires only lightweight adaptation. With only 8,000 gameplay samples, 10,000 LoRA optimization steps, and 0.199\% trainable parameters, \method achieves effective character and camera control while preserving strong generation quality. It also generalizes to unseen scenarios. These results show that the control capabilities emerging in large video generators can be efficiently transformed into interactive world control.
}

\maketitle

\begingroup
\renewcommand{\thefootnote}{\ensuremath{*}}
\footnotetext[0]{Corresponding authors.}
\renewcommand{\thefootnote}{\ensuremath{\clubsuit}}
\footnotetext[0]{This work was completed during a research internship at Tencent, supervised by Yeying Jin.}
\renewcommand{\thefootnote}{\ensuremath{\diamondsuit}}
\footnotetext[0]{Project leader.}
\endgroup

\section{Introduction}
\label{sec:intro}

Language is becoming more than a way to describe generated worlds. As video generators grow more capable, language is also beginning to shape how those worlds evolve---how agents move, how cameras behave, and how scenes change over time~\cite{minimax,genie,Wan}. In this sense, language is emerging as a high-level abstraction layer for visual dynamics. This shift is especially relevant to world models, which learn how environments evolve under actions and connect intelligent agents with their surroundings~\cite{worldmodels,planet,dreamer,dreamerv2}. Meanwhile, advances in diffusion transformers and large-scale video generation have made pretrained video models a natural foundation for visually rich world simulation~\cite{dit,svd,latte,cogvideox,videocrafter2,opensora,hunyuanvideo,Wan,minimax}. Such models already support increasingly rich forms of interactive content, game simulation, and learned real-world simulation~\cite{genie,unisim,gamegenx,thematrix,matrixgame1,worldplay}.

Yet generation is not control. A pretrained video model does not, by itself, expose the precise action interface required for interactive world modeling. Existing systems therefore introduce new control pathways through discrete action embeddings, conditional modules, camera geometry, or direct backbone adaptation~\cite{gamegan,GameNGen,diamond,gamefactory,Lingbot-world,lingbotworld2,matrixgame2,matrixgame3,abotworld}. These pathways are typically learned from temporally aligned action-video trajectories. They require additional supervision and adaptation on top of the pretrained generator. They can also increase computation and storage, and extensive adaptation may disturb capabilities acquired during pretraining. More fundamentally, this paradigm treats control as something that must be learned largely on top of the pretrained model.

\begin{figure}[!ht]
\centering
\includegraphics[width=\textwidth]{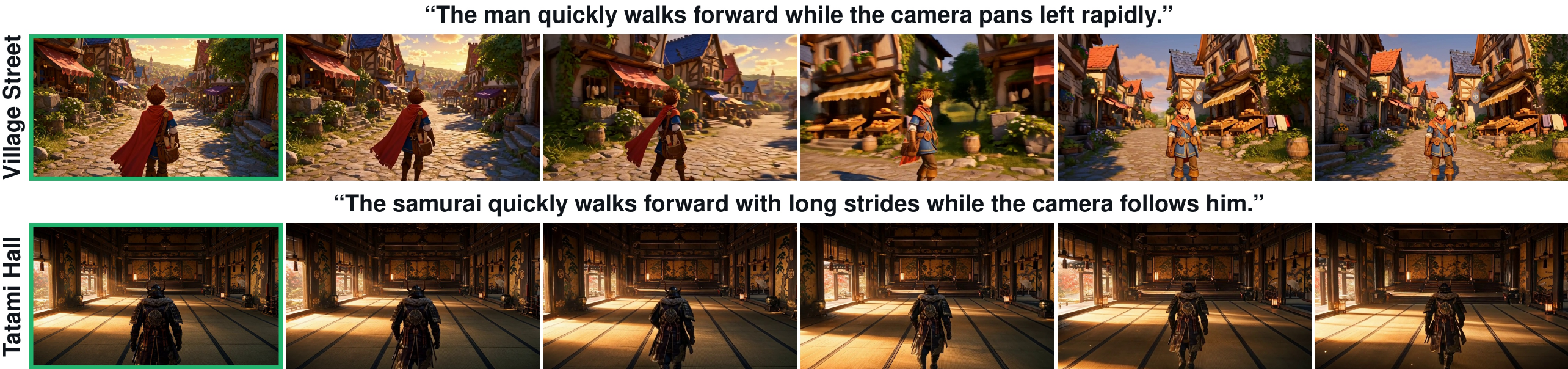}
\caption{
Vanilla MiniMax-H3 already exhibits coarse zero-shot control through textual motion instructions.
}
\label{fig_intro_prior}
\end{figure}

But large video generators may already contain part of this bridge. Modern video models learn rich representations of objects, agents, motion, and interaction from large-scale video data~\cite{svd,cogvideox,hunyuanvideo,Wan,minimax}. MiniMax-H3, for example, can already follow coarse textual instructions for character and camera motion~\cite{minimax}. As shown in Figure~\ref{fig_intro_prior}, such instructions produce roughly correct responses with realistic visual dynamics, even without action-conditioned training. This observation is central to our work: \emph{language already acts as a coarse control interface for strong text-to-video models}.

Based on this observation, we propose \method, an efficient framework that turns the 33B-parameter MiniMax-H3 into an interactive world model. Our idea is simple. Instead of learning a new control representation from scratch, we express control in the language space the model already understands. Accordingly, \method converts character and camera actions into compositional textual instructions and injects them through the native text pathway of MiniMax-H3. To make these controls precise over time, we align each instruction with its corresponding video latent interval. We further introduce temporal attention routing to preserve this alignment and prevent control leakage across time. As a result, \method directly reuses the pretrained language and generative capabilities of MiniMax-H3 without introducing a dedicated action-specific control module. This design requires only lightweight LoRA adaptation~\cite{hu2022lora,difffit}.

Surprisingly, this lightweight adaptation is sufficient to produce effective interactive control. With only 8,000 gameplay samples and 0.199\% trainable parameters, \method learns precise character and camera control while preserving the generative capabilities of the pretrained model. More importantly, the resulting control transfers beyond the training distribution, including unseen action compositions and distinct initial observations. Together, these results suggest that much of the structure needed for interactive control can already exist inside a sufficiently capable video generator. By grounding low-level actions into its native language interface, \method provides a direct path from pretrained video generation to interactive world modeling.

In summary, our contributions are as follows.
\begin{itemize}[leftmargin=*,topsep=3pt,itemsep=2pt]
\item We show that large video generators already support coarse language-based control, providing a strong basis for interactive world modeling.
\item We introduce \method, which turns the language understanding of large video models into world control. It represents character and camera actions as textual instructions and uses temporal attention routing for precise control, without dedicated action-specific modules.
\item With only 8,000 gameplay samples and 0.199\% trainable parameters, \method achieves effective control while preserving generation quality and generalizing to unseen actions and visual scenarios.
\end{itemize}

\section{Related Work}
\label{sec:related}

\subsection{Action-Conditioned Video World Models}
Interactive world models generate visual futures in response to user actions. Earlier neural and diffusion game simulators condition visual dynamics on recorded actions~\cite{gamegan,GameNGen,diamond}, while Genie learns a latent action interface from video~\cite{genie}. Recent systems extend stronger video generators toward open worlds, device control, causal rollout, persistent history, and real-time inference~\cite{gamegenx,thematrix,matrixgame1,gamefactory,oasis,Lingbot-world,lingbotworld2,matrixgame2,matrixgame3,worldplay,abotworld}. Related models study reactive characters, cross-game interaction, and explicit state-aware behavior~\cite{reactivegwm,scope,stateplay,worldmind}. These methods typically establish control through learned embeddings, feature modulation, camera geometry, or dedicated modules. Such interfaces require additional control learning and may limit robustness and generalization~\cite{li2026badwam,shen2026badworld}. In contrast, \method injects fine-grained controls directly into the textual prompt, allowing them to be interpreted through the language understanding already acquired during video pretraining. This directly reuses the pretrained semantic prior, requires adapting only 0.199\% of the parameters~\cite{minimax,hu2022lora,difffit}, and preserves strong generalization to unseen scenarios.

\subsection{Language Actions and Temporal Grounding}
Recent work uses language to express richer interventions and world-model actions~\cite{unisim,gamecraft2,yume15,actworld,incantation,helloworld}. Incantation is most closely related in representing actions with per-latent-frame natural language and local text cross-attention~\cite{incantation}. Beyond world models, controllable video generation uses trajectories, camera poses, and compositional conditions to direct object and viewpoint motion~\cite{videocomposer,motionctrl,cameractrl,directavideo,dragnuwa}. These works establish semantic conditioning as a useful control interface. \method considers a different architectural setting in which semantic conditions and video latents coexist in MiniMax-H3's packed single-stream self-attention. The model jointly denoises the full future horizon without a separate text cross-attention layer, so our focus is the temporal grounding of language actions within this bidirectional sequence.

\section{Method}
\label{sec_method}

\subsection{Overview}
\label{sec_method_overview}

\begin{figure*}[t]
    \centering
    \includegraphics[width=\textwidth]{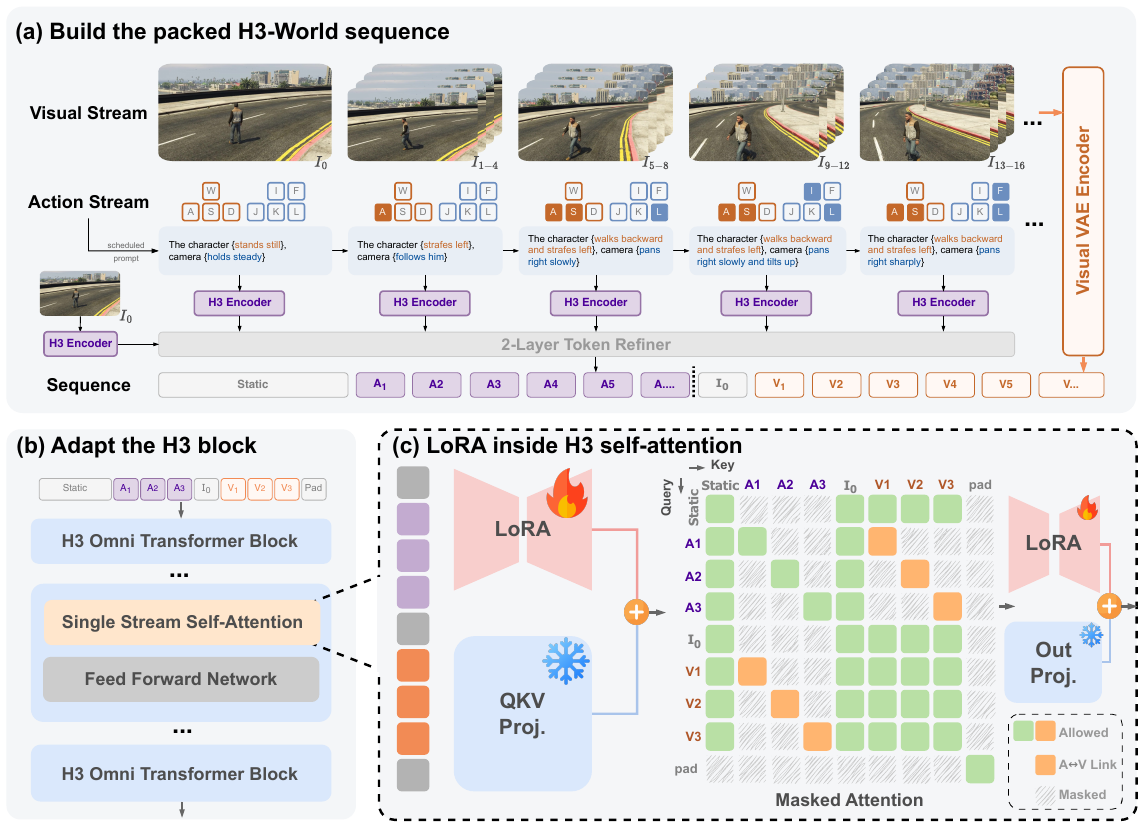}
    \caption{
        Overview of \method.
        (a) Latent-aligned action prompts are independently encoded and packed with the static semantic condition, initial observation, and video latents.
        (b) MiniMax-H3 processes the packed sequence through single-stream self-attention.
        (c) LoRA adapts the attention projections under single-egress routing, which connects each action span directly to its matched video latent and retains bidirectional attention among video latents.
    }
    \label{fig_h3world_method}
\end{figure*}

Given an initial observation $I_0$, a static semantic condition $s$, and a scheduled action sequence $\mathbf{a}_{1:K}$, \method generates the corresponding future video latents $\mathbf{V}_{1:K}$. We build on MiniMax-H3~\cite{minimax}, a pretrained bidirectional audio-video foundation model that jointly denoises the complete future horizon. Its pretrained parameters $\theta$ remain frozen, and $\phi$ denotes the lightweight adaptation parameters learned for action control. The conditional generation process is written as
\begin{equation}
    \widehat{\mathbf{V}}_{1:K}
    \sim
    p_{\theta,\phi}
    \left(
        \mathbf{V}_{1:K}
        \mid
        I_0,
        s,
        \mathbf{a}_{1:K}
    \right).
    \label{eq_task}
\end{equation}
MiniMax-H3 processes text, image, audio, and video tokens in a shared sequence. We retain its native audio stream and focus the following notation on visual action control.

As shown in Fig.~\ref{fig_h3world_method}, \method contains three components. The semantic action interface converts character and camera controls into compositional textual instructions. Latent-aligned temporal binding associates each instruction with a video latent interval. Single-egress routing maintains this association throughout the H3 backbone, while LoRA learns the corresponding action-conditioned visual dynamics.

\subsection{Semantic Action Interface}
\label{sec_semantic_action}

External controls are represented as discrete control states. Each state contains eight recorded character and camera keys and one binary camera-speed flag. During training, the speed flag is derived from the estimated camera yaw rate. At inference, it is specified directly by the user. Each native H3 video latent represents a short interval of RGB frames. We aggregate the recorded key states within that interval into one latent-level state, marking a key as active when it occurs in any frame of the interval. Opposing keys are cancelled before prompt construction.

These control states provide compact and temporally precise commands, while their symbols carry limited information about the corresponding visual transitions. We expose the compositional structure of the control space by separating character control from camera control. For the $k$-th video latent interval, the action is written as
\begin{equation}
    \mathbf{a}_k
    =
    \left(
        u_k,
        c_k
    \right),
    \qquad
    u_k \in \mathcal{U},
    \qquad
    c_k \in \mathcal{C},
    \label{eq_action_factorization}
\end{equation}
where $\mathcal{U}$ contains the character-control commands and $\mathcal{C}$ contains the camera-control commands. The resulting action space contains nine character clauses and sixteen camera clauses.

We map each control pair to a short textual instruction
\begin{equation}
    p_k
    =
    \mathcal{T}_{\mathrm{char}}(u_k)
    \,\Vert\,
    \mathcal{T}_{\mathrm{cam}}(c_k),
    \label{eq_action_prompt}
\end{equation}
where $\Vert$ denotes clause concatenation. For example, a backward-left character command combined with a slow right-pan camera command produces the prompt ``the man walks backward and strafes left, camera pans right slowly.'' The character and camera clauses follow a shared grammatical structure across the dataset. This representation places external controls in the native text-conditioning space of H3 and preserves the factorization of the original action space.

\begin{figure*}[t]
    \centering
    \includegraphics[width=\textwidth]{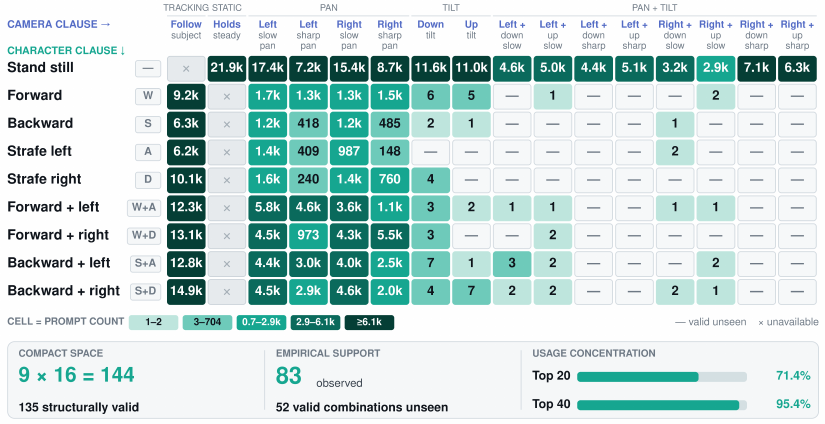}
    \caption{
        Training coverage of the action space.
        Of 135 valid character--camera pairs, 83 occur in 291,264 prompts and 52 remain unseen.
        The top 20 and 40 pairs account for 71.4\% and 95.4\% of prompts.
    }
    \label{fig_action_space}
\end{figure*}

The action space is compact, but the training data cover only a sparse and highly imbalanced subset of the valid combinations. Figure~\ref{fig_action_space} summarizes the joint distribution of character and camera clauses in the training set. The Cartesian product contains $9 \times 16 = 144$ combinations, of which 135 are structurally valid under our control specification. The training data covers 83 valid combinations and leaves 52 valid combinations unseen. The observed support therefore satisfies
\begin{equation}
    \mathcal{A}_{\mathrm{train}}
    \subsetneq
    \mathcal{A}_{\mathrm{valid}}
    \subsetneq
    \mathcal{U}\times\mathcal{C}.
    \label{eq_action_support}
\end{equation}
The distribution is concentrated among a small number of frequent pairs. The 20 most frequent combinations account for 71.4\% of all action prompts, and the 40 most frequent combinations account for 95.4\%. This distribution provides a natural setting for evaluating compositional control. A joint command may be absent from the training set even when its character and camera clauses occur in other combinations.

The semantic action interface specifies the requested visual transition. Time-varying interaction also requires a correspondence between each instruction and a particular portion of the generated video. We establish this correspondence in the H3 token sequence.

\subsection{Latent-Aligned Temporal Binding}
\label{sec_temporal_binding}

A video-level prompt provides a shared semantic condition for an entire clip and offers limited temporal resolution when the requested action changes within the generation horizon. We assign an independent action prompt to every video latent interval. Each prompt $p_k$ is encoded by the shared H3 encoder $\mathcal{E}$ and processed by a shared two-layer token refiner $\mathcal{R}$
\begin{equation}
    \mathbf{A}_k
    =
    \mathcal{R}
    \left(
        \mathcal{E}(p_k)
    \right).
    \label{eq_action_encoding}
\end{equation}
The token refiner uses block-diagonal attention over the scheduled prompts. Tokens within the same action span communicate bidirectionally, and different action spans are processed as separate sequences. This layout uses a common representation space for all actions and preserves the temporal identity of each instruction before it enters the video backbone.

The initial observation follows two complementary encoding paths. The multimodal H3 encoder jointly processes the static semantic condition $s$ and the initial observation $I_0$ to produce static semantic tokens $\mathbf{S}$. The visual VAE maps $I_0$ to a first-frame condition $\mathbf{C}_0$ that preserves fine-grained appearance. The target video is encoded into latent intervals $\mathbf{V}_{1:K}$. During training, these intervals contain the noised target latents used by the native H3 denoising objective.

We pack the visual conditions and generation targets into a single sequence
\begin{equation}
    \mathbf{X}
    =
    \left[
        \mathbf{S};
        \mathbf{A}_1;
        \ldots;
        \mathbf{A}_K;
        \mathbf{C}_0;
        \mathbf{V}_1;
        \ldots;
        \mathbf{V}_K;
        \mathbf{P}
    \right],
    \label{eq_packed_sequence}
\end{equation}
where $\mathbf{P}$ denotes masked padding tokens. Each action span $\mathbf{A}_k$ is paired with the $k$-th target video latent under the native temporal partition of the H3 VAE. A video latent interval may represent several RGB frames after VAE decoding.

We assign mirrored temporal positions to the action spans. Let $\tau(\cdot)$ denote the temporal coordinate used by the H3 positional encoding. The position of each action span follows the temporal order of its matched video latent and remains in the text-side positional range. Every matched pair therefore has the same relative temporal offset
\begin{equation}
    \tau(\mathbf{A}_k)
    =
    \tau(\mathbf{V}_k)
    -
    \Delta,
    \qquad
    \Delta > 0.
    \label{eq_mirrored_position}
\end{equation}
This construction preserves the text-before-video ordering of pretrained H3 and provides a consistent temporal alignment cue for every action and latent pair.

Temporal alignment alone does not restrict information flow in bidirectional self-attention. An action span could still communicate directly with unmatched video latents. We therefore introduce a routing structure that constrains the direct path from each action span to the video stream.

\subsection{Single-Egress Routing and Low-Rank Adaptation}
\label{sec_single_egress}

We construct a deterministic single-egress routing mask for each packed sequence. When an action span $\mathbf{A}_k$ serves as an attention key, it can be read by tokens in the same action span and by its matched video latent $\mathbf{V}_k$. Static tokens, first-frame condition tokens, native audio tokens, other action spans, and unmatched video latents cannot read $\mathbf{A}_k$. When $\mathbf{A}_k$ serves as an attention query, it retains access to the static context, first-frame condition, native audio context, its own tokens, and the matched video latent. Access to other action spans and unmatched video latents is masked.

All video latent spans retain the original bidirectional attention pattern of H3. The visual effect associated with $\mathbf{A}_k$ first enters the video stream through $\mathbf{V}_k$ and can subsequently propagate through video-to-video attention. This structure preserves full-horizon information exchange for motion continuity and scene consistency. It also gives every scheduled action a unique direct entry point into the visual stream. We refer to this property as single-egress action routing.

LoRA~\cite{hu2022lora} learns the action-conditioned transformations along the permitted routes. We apply low-rank updates to the QKV and output projections in the H3 self-attention blocks and to the two-layer token refiner. The backbone, H3 encoder, visual VAE, and all remaining components remain frozen. The routing mask and span partition introduce no learnable parameters, and training retains the native H3 denoising objective.

\section{Experiments}
\label{sec:experiments}

\subsection{Experimental Setup}
\label{sec_experimental_setup}

\paragraph{Data.}
We construct the training and evaluation sets from ABot-World-Explorer-500h~\cite{abotworld}.
The training set contains 7,872 gameplay clips, and a separate set of 128 clips is held out for evaluation.
Each clip contains 124 frames at 24 fps with a resolution of $832 \times 480$.
Following the temporal partition described in Section~\ref{sec_temporal_binding}, every clip provides 37 latent-aligned action prompts.

\paragraph{Training and inference.}
We train the rank-32 LoRA parameters for 10,000 optimization steps with a learning rate of $1\times10^{-4}$.
At inference, we generate 124 frames with 50 denoising steps.
The pretrained H3 comparisons use the same generation settings with the LoRA updates disabled.

\paragraph{Evaluation protocol.}
We evaluate action control with two paired protocols. For controlled interventions, the initial observation, generation seed, and sampling configuration are fixed while only the scheduled action changes. For held-out clips, we condition \method on the initial observation and recorded action sequence, then compare the generated and ground-truth videos by their motion patterns.

Qualitative figures use uniformly sampled frames.
The green border marks the initial observation, and the keyboard overlay shows the scheduled control state.
Pressed keys are highlighted in orange.
When a quantitative diagnostic is required, we estimate dense optical flow using the Farneback method~\cite{farneback2003two} and accumulate the mean horizontal flow over each video.
Positive and negative values indicate leftward and rightward camera motion, respectively.

\subsection{Pretrained Action Prior and Adaptation}
\label{sec_pretrained_action_prior}

\begin{figure*}[!htbp]
    \centering
    \includegraphics[width=\textwidth]{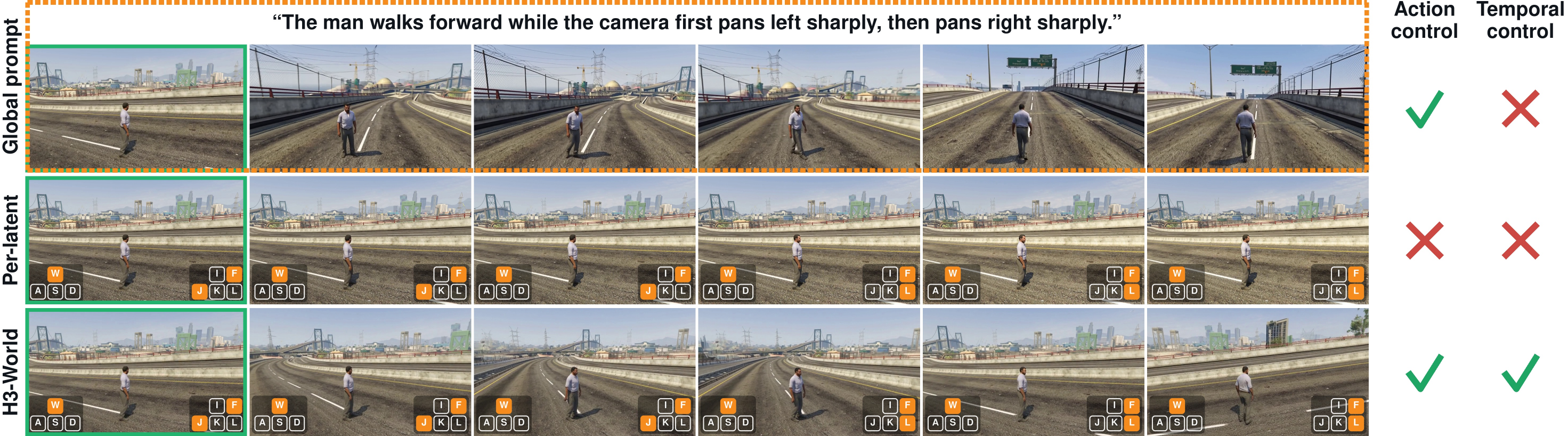}
    \caption{
        Action response before and after adaptation.
        The scheduled camera pan switches from left to right after latent 15.
        Check marks and crosses summarize action control and temporal control for each condition.
    }
    \label{fig_pretrained_action_prior}
\end{figure*}

We examine whether adaptation enables temporally specified control beyond the coarse motion response already available in pretrained H3.
Figure~\ref{fig_pretrained_action_prior} uses a controlled camera schedule that pans left sharply for the first 15 temporal latents and right sharply for the remaining 22.
The switch coincides with a latent block boundary.
This schedule requires the model to follow both directions and assign each one to its requested interval, while the reversal within one clip controls for scene drift.
We compare three settings: (1) frozen H3 with one global motion instruction appended to the scene prompt; (2) the complete per-latent action interface with every LoRA update set to zero; and (3) the trained \method checkpoint.
We report cumulative horizontal flow separately for the frames before and after the switch.


The three conditions exhibit distinct responses.
Global prompting produces cumulative horizontal flow of $0.0$ before the switch and $-17.3$ afterward.
Frozen H3 therefore shows a small rightward response in the second half, while the scheduled leftward motion is absent.
The zero-LoRA per-latent interface remains nearly static, with flows of $-0.1$ and $0.0$ and mean absolute horizontal flow of $0.003$.
In contrast, \method produces $+52.7$ before the switch and $-106.0$ afterward, following both scheduled directions.
Reversing the instruction order gives the same pattern: global prompting yields $-11.9$ and $+24.1$, the zero-LoRA interface remains unresponsive, and \method yields $-58.7$ and $+121.0$.

The constant-action control clarifies the role of adaptation.
With one camera direction throughout the clip, global prompting and \method yield nearly identical directional separation, $301.8$ and $300.5$.
Frozen H3 can therefore respond to a coarse motion instruction.
For a changing schedule, however, its global text representation is shared across the full horizon and does not bind each direction to a latent interval.
The zero-LoRA condition shows that supplying span-specific instructions alone is insufficient.
LoRA adaptation enables the backbone to use these temporal bindings and follow both parts of the schedule.

\subsection{Action Controllability}
\label{sec_action_controllability}

\begin{figure*}[!htbp]
    \centering
    \includegraphics[width=\textwidth]{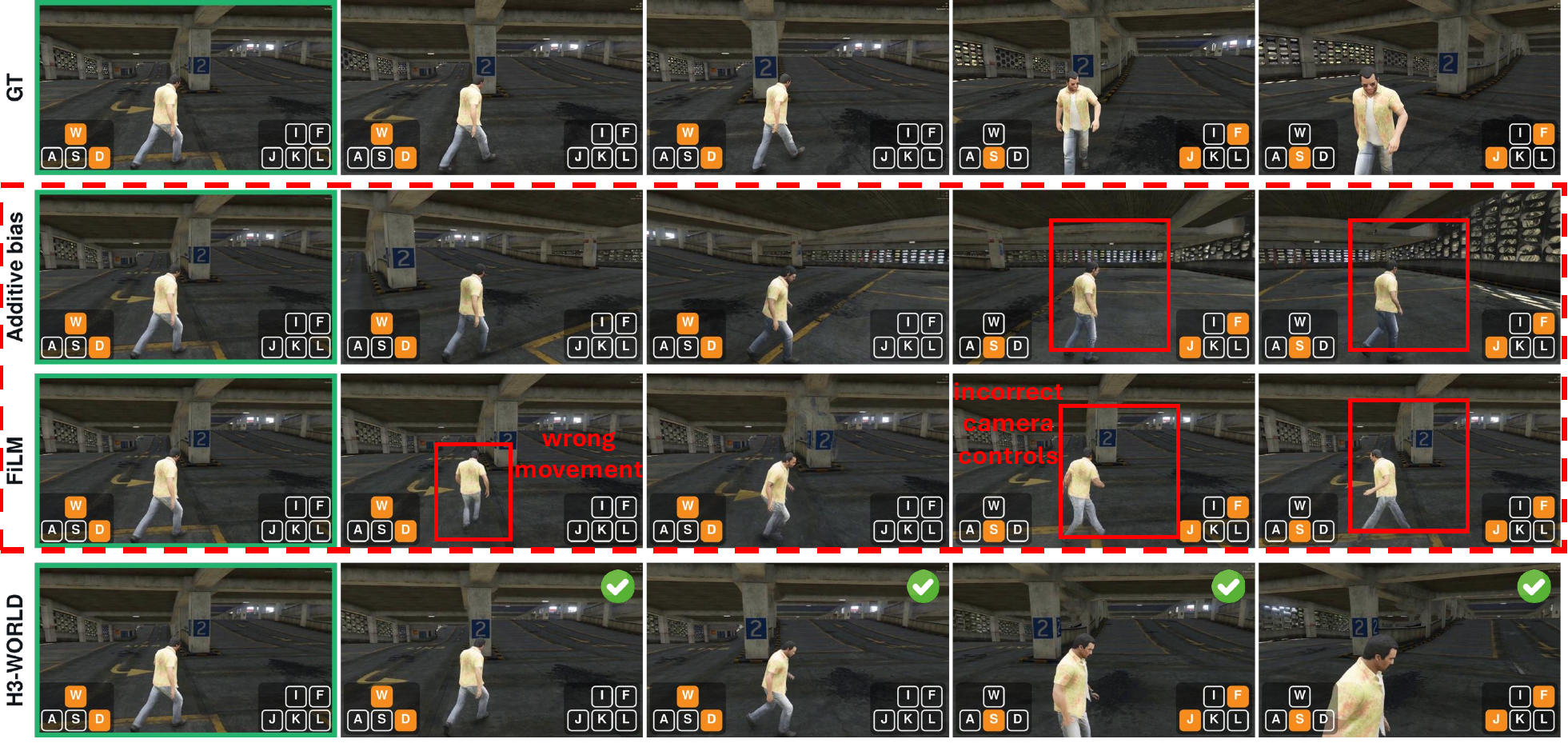}
    \caption{
        Action-conditioning interfaces on a held-out clip.
        GT provides the recorded reference motion.
        Direct additive-bias and FiLM conditioning yield limited or inconsistent responses to the recorded controls, while text-based \method produces coordinated character and camera changes.
    }
    \label{fig_feature_injection}
\end{figure*}

\paragraph{Direct action conditioning.}
We first compare the text-based action interface of \method with two direct action-conditioning variants.
Each variant encodes the latent-level keyboard state as a learned vector and applies it to the corresponding video-token features.
The additive-bias variant follows the action-conditioning mechanism used in ReactiveGWM~\cite{reactivegwm}, which projects the action state and adds it to the video representation.
The FiLM variant instead applies feature-wise scale and shift after AdaLN~\cite{perez2018film}.
Figure~\ref{fig_feature_injection} compares the three interfaces with the ground-truth video from the same held-out clip and recorded action sequence.
The additive and FiLM variants produce weak or inconsistent changes as the recorded controls vary.
In contrast, \method produces coordinated character and camera changes throughout the sequence.
This diagnostic supports grounding actions in the pretrained text pathway for action control.

\begin{figure*}[!ht]
    \centering
    \includegraphics[width=0.97\textwidth]{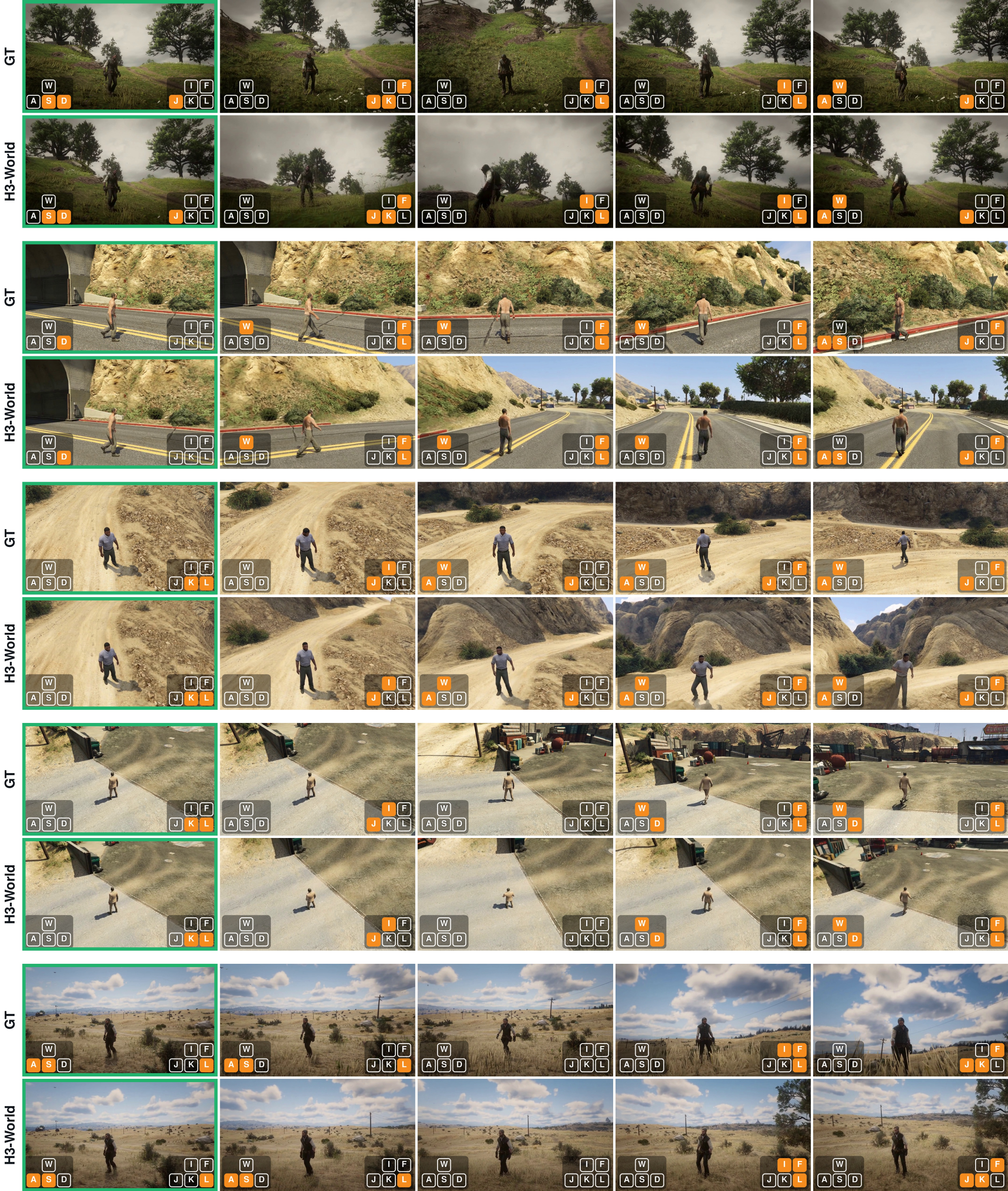}
    \caption{
        Following recorded controls on held-out clips.
        Ground-truth videos appear above the corresponding \method generations produced from the same initial observation and action sequence.
    }
    \label{fig_gt_action_control}
\end{figure*}

\paragraph{Following recorded controls.}
We next evaluate the learned text-based interface on held-out gameplay clips.
Figure~\ref{fig_gt_action_control} pairs each clip with a \method generation conditioned on its first frame and recorded action sequence.
Each clip admits multiple plausible futures, so we assess whether the generation exhibits the recorded character and camera motion while retaining the initial scene content.
Across the five clips, \method exhibits the main character displacement and viewpoint changes indicated by the recorded controls while preserving the scene layout and subject appearance.
This indicates that the learned interface transfers recorded action semantics to held-out gameplay observations.

\begin{figure*}[!htbp]
    \centering
    \includegraphics[width=\textwidth]{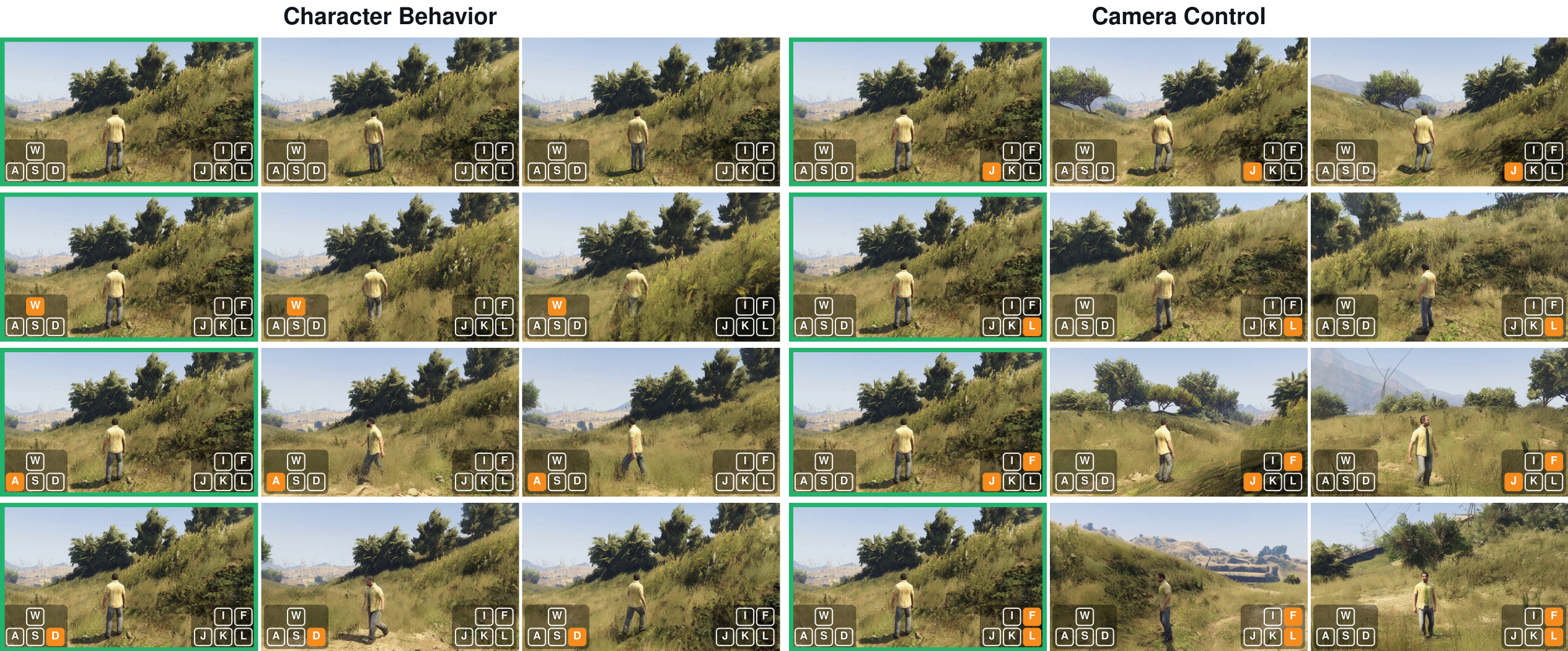}
    \caption{
        Controlled action comparison.
        The initial observation, seed, and sampling configuration are fixed across rows.
        Changing the action produces distinct character and camera motion, including stronger fast pans.
    }
    \label{fig_action_intervention}
\end{figure*}

\paragraph{Controlled action intervention.}
We then fix the initial observation and generation seed and change only the action instruction.
Figure~\ref{fig_action_intervention} includes stationary and forward motion, left and right strafing, and slow and fast camera pans in both directions.
Opposite strafe and pan commands produce distinct lateral evolution.
The fast camera commands also produce visibly larger changes than their slow counterparts.
The paired construction shows that the generated differences arise from the requested action.

\subsection{Generalization}
\label{sec_generalization}

\paragraph{Compositional action generalization.}

\begin{figure*}[!htbp]
    \centering
    \includegraphics[width=\textwidth]{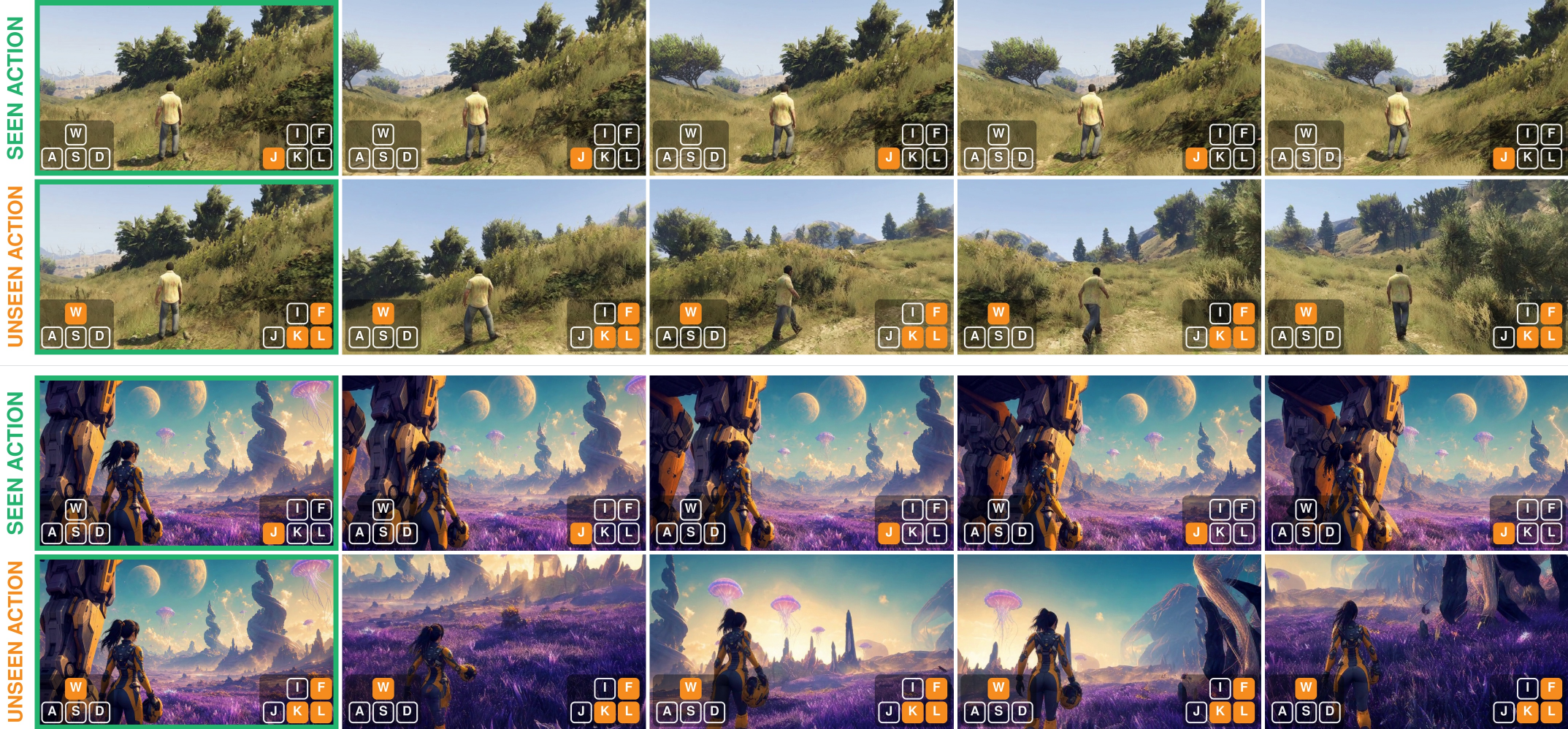}
    \caption{
        Compositional action generalization.
        Each group contrasts a seen action with an unseen composition of observed character and camera clauses on held-out gameplay (top) and out-of-distribution (bottom) observations.
    }
    \label{fig_action_composition}
\end{figure*}

Of the 135 structurally valid action combinations, 83 occur in training.
We evaluate an unseen pair whose character and camera clauses each occur in other training pairs, but never together.
This setting tests whether the semantic interface composes familiar control primitives beyond the observed joint support. Figure~\ref{fig_action_composition} contrasts this unseen forward-motion and camera pan--tilt command with a seen reference on held-out gameplay and out-of-distribution observations.
\method follows both the character and camera components of the unseen command in both settings while preserving the scene layout and subject appearance.
This result shows that compositional action control extends beyond the joint combinations observed during training.

\begin{figure*}[!htbp]
    \centering
    \includegraphics[width=\textwidth,height=0.86\textheight,keepaspectratio]{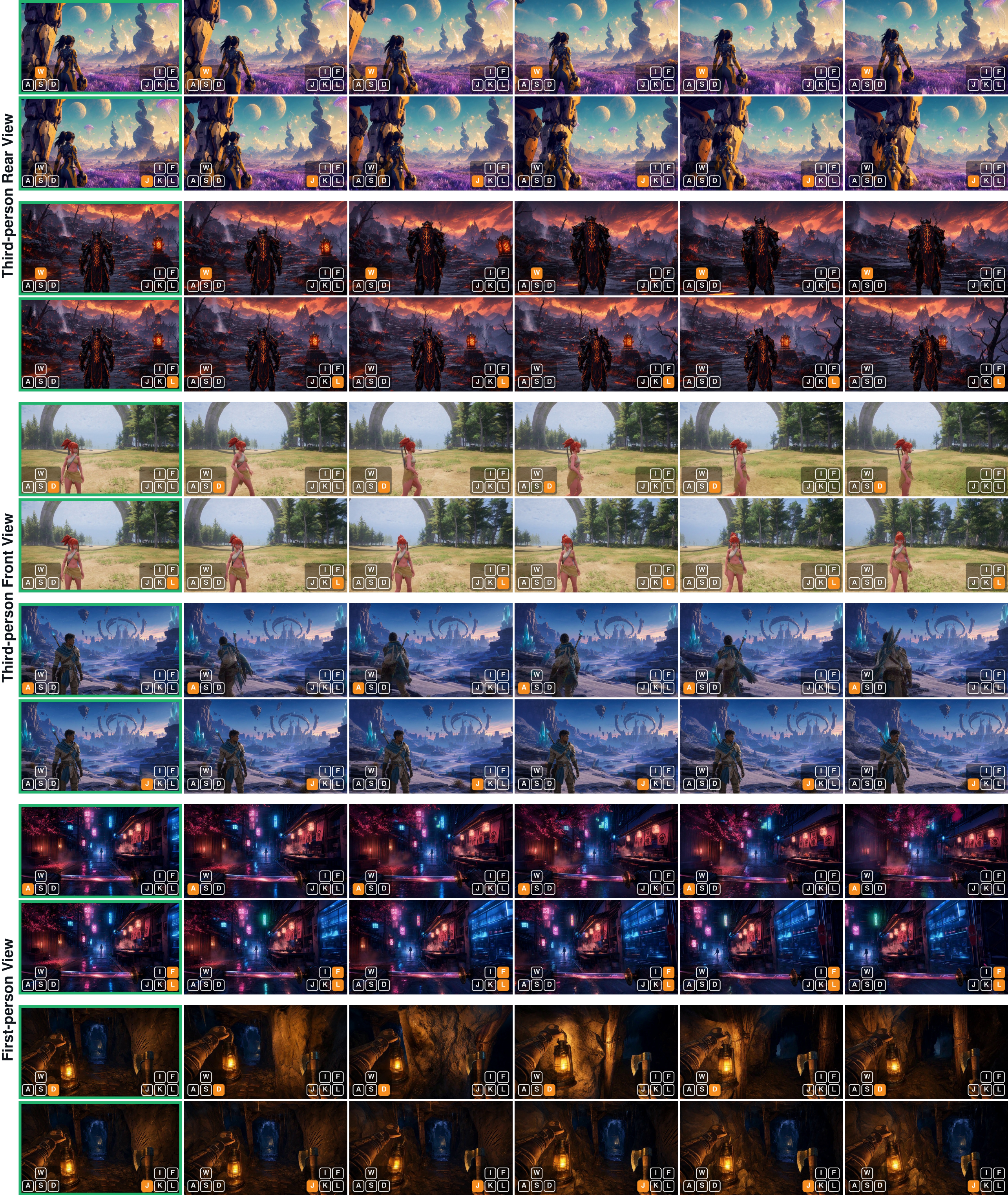}
    \caption{
        Visual generalization to diverse initial observations.
        Each scene is evaluated with one character-control command and one camera-control command using the same learned action interface.
    }
    \label{fig_visual_generalization}
\end{figure*}

\paragraph{Visual generalization.}
We further evaluate the learned action interface on six initial observations that differ substantially from the gameplay training set.
The scenes cover third-person and first-person viewpoints, indoor and outdoor environments, fantasy and science-fiction content, and diverse rendering styles.

As shown in Figure~\ref{fig_visual_generalization}, the same interface produces the requested character or camera response across these visual domains.
The generations preserve the scene layout, subject identity, and visual style of each initial observation.
These results support the use of lightweight adaptation for retaining the broad generative capabilities of pretrained H3 while learning interactive control from a small gameplay dataset.

\section{Conclusion}
\label{sec:conclusion}

We present \method, an efficient adaptation of MiniMax-H3 for interactive world modeling. \method expresses character and camera commands as compositional text instructions, aligns them with video latent intervals, preserves the alignment through single-egress routing, and learns the associated visual dynamics through low-rank adaptation. The results show that this design supports recorded and intervened control, generalizes to unseen action pairs and visually distinct initial observations, and reuses the control capabilities already present in a pretrained video generator. These findings show that temporally grounded language conditioning can provide an effective route from pretrained video generation to low-level world control.

However, there are still few limitations in this work. Firstly, our current study focuses on short-horizon generation, with compositional generalization and visual transfer evaluated mainly through representative examples; more systematic evaluation across action combinations, scenes, and random seeds is needed to quantify control reliability. Secondly, the current model generates fixed-length segments and does not yet provide persistent world state, real-time interaction, planning, or policy learning. Extending the interface in these directions will be important for interactive world models that operate over longer horizons.

\section*{Acknowledgements}

We would like to express our sincere gratitude to Ruidong Wang and Murphy Zhao for their invaluable support throughout this project.

\clearpage
\bibliography{references}

\end{document}